\documentclass{article}

\PassOptionsToPackage{numbers}{natbib}

\usepackage[final]{neurips_2024}

\usepackage[utf8]{inputenc}
\usepackage[T1]{fontenc}
\usepackage{hyperref}
\usepackage{url}
\usepackage{booktabs}
\usepackage{amsfonts}
\usepackage{amsmath}
\usepackage{amsthm}
\usepackage{amssymb}
\usepackage{nicefrac}
\usepackage{microtype}
\usepackage{xcolor}
\usepackage{graphicx}
\usepackage{subcaption}
\usepackage{multirow}

\newtheorem{theorem}{Theorem}

\newtheorem{definition}[theorem]{Definition}

\title{Detecting AI Hallucinations in Finance: An Information-Theoretic Method Cuts Hallucination Rate by 92\%}

\author{
  \large Mainak Singha\\[0.3cm]
  Astrophysics Science Division, NASA, Goddard Space Flight Center,\\
  8800 Greenbelt Road, MD 20771\\
  Department of Physics, The Catholic University of America, Washington, DC 20064\\[0.3cm]
  \texttt{mainak.singha@nasa.gov, singham@cua.edu}
}

\begin{document}

\maketitle

\begin{abstract}
Large language models (LLMs) produce fluent but unsupported answers---\emph{hallucinations}---limiting safe deployment in high-stakes domains. We propose ECLIPSE, a framework that treats hallucination as a mismatch between a model's \emph{semantic entropy} and the \emph{capacity} of available evidence. We combine entropy estimation via multi-sample clustering with a novel \emph{perplexity decomposition} that measures how models use retrieved evidence. We prove that under mild conditions, the resulting entropy--capacity objective is strictly convex with a unique stable optimum. We evaluate on a controlled financial question answering dataset with GPT-3.5-turbo ($n=200$ balanced samples with synthetic hallucinations), where ECLIPSE achieves ROC AUC of 0.89 and average precision of 0.90, substantially outperforming a semantic entropy-only baseline (AUC 0.50). A controlled ablation with Claude-3-Haiku, which lacks token-level log probabilities, shows AUC dropping to 0.59 with coefficient magnitudes decreasing by 95\%---demonstrating that ECLIPSE is a \emph{logprob-native} mechanism whose effectiveness depends on calibrated token-level uncertainties. The perplexity decomposition features exhibit the largest learned coefficients, confirming that evidence utilization is central to hallucination detection. We position this work as a controlled mechanism study; broader validation across domains and naturally occurring hallucinations remains future work.
\end{abstract}

\section{Introduction}

Large language models have become integral to decision support in finance \cite{wu2023bloomberggpt}, healthcare \cite{singhal2023large}, and law \cite{cui2023chatlaw}. However, their tendency to produce confident but factually incorrect statements---commonly termed \emph{hallucinations}---poses significant risks in these high-stakes domains \cite{ji2023survey, huang2023survey}. Hallucinations are particularly dangerous because models often express them with high apparent certainty, making them difficult to detect through simple confidence thresholds \cite{xiong2024can}.

Existing hallucination detection methods primarily measure model uncertainty through semantic entropy \cite{kuhn2023semantic} or self-consistency checks \cite{manakul2023selfcheckgpt}. However, uncertainty alone provides an incomplete picture: a model may exhibit honest uncertainty when evidence is weak, or dangerous overconfidence when evidence is strong but ignored. We argue that hallucination risk depends fundamentally on the \emph{relationship} between two quantities: the model's expressed certainty and the quality of available evidence.

We propose ECLIPSE (\textbf{E}ntropy--\textbf{C}apacity \textbf{L}ogprob-Native \textbf{I}nference for \textbf{P}redicting \textbf{S}purious \textbf{E}missions), a framework that makes this entropy--capacity trade-off explicit. We call a method \emph{logprob-native} if its core signal relies directly on token-level log probabilities and degrades substantially when those probabilities are replaced by uninformative proxies. Unlike semantic entropy alone, which cannot distinguish honest uncertainty from evidence-ignoring behavior, ECLIPSE decomposes perplexity to measure \emph{how} models use evidence. Unlike Semantic Entropy Probes (SEPs), which require white-box access to hidden states, ECLIPSE achieves comparable detection performance using only API-accessible log probabilities. Unlike SelfCheckGPT, which relies on self-consistency across generations, ECLIPSE directly measures the evidence--answer relationship through perplexity decomposition.

Our contributions are:

\begin{enumerate}
    \item \textbf{Theoretical framework}: We introduce a joint objective over semantic entropy $H$ and evidence capacity $C$ and prove it is strictly convex under mild conditions (Theorem~\ref{thm:stability}), providing a principled foundation for hallucination risk modeling and, in future work, control.
    
    \item \textbf{Perplexity decomposition}: We extract features ($L_Q$, $L_{QE}$, $\Delta L$) that decompose \emph{how} models use evidence, enabling fine-grained detection of evidence-ignoring behavior.
    
    \item \textbf{Empirical validation}: On our controlled financial QA dataset ($n=200$), ECLIPSE achieves 0.89 AUC using only API access (Figure~\ref{fig:roc}), indicating that logprob-native, grey-box detectors can reach strong performance without hidden-state access.
    
    \item \textbf{Mechanism validation}: We show through controlled ablation that ECLIPSE is logprob-native: without real log probabilities, performance drops from 0.89 to 0.59 AUC, only modestly above chance (Figure~\ref{fig:gpt_claude}).
\end{enumerate}

\section{Related Work}

\paragraph{Hallucination in language models.}
The phenomenon of LLM hallucination has received substantial attention \cite{ji2023survey, zhang2023siren, huang2023survey}. Hallucinations manifest in various forms: factual errors \cite{min2023factscore}, entity confusion \cite{dziri2022origin}, and fabricated citations \cite{liu2023evaluating}. Maynez et al.\ \cite{maynez2020faithfulness} categorize hallucinations in abstractive summarization as intrinsic (contradicting source) or extrinsic (unverifiable from source). Our work focuses on the latter in retrieval-augmented settings.

\paragraph{Uncertainty quantification in LLMs.}
Gal and Ghahramani \cite{gal2016dropout} established dropout-based uncertainty estimation for neural networks. For LLMs, Kadavath et al.\ \cite{kadavath2022language} demonstrated that models can partially assess their own uncertainty but degrade out-of-distribution. Kuhn et al.\ \cite{kuhn2023semantic} introduced semantic entropy, which clusters sampled outputs by meaning and computes entropy over clusters. Lin et al.\ \cite{lin2024generating} extended this with conformal prediction for calibrated uncertainty sets. We build on semantic entropy but augment it with evidence capacity to distinguish honest uncertainty from hallucination.

\paragraph{Hallucination detection methods.}
SelfCheckGPT \cite{manakul2023selfcheckgpt} samples multiple responses and measures consistency, achieving strong results on biography generation. Semantic Entropy Probes (SEPs) \cite{kossen2024semantic} train linear classifiers on hidden states, reaching 0.85--0.90 AUC but requiring white-box model access. SAPLMA \cite{azaria2023internal} similarly uses hidden state classifiers. Min et al.\ \cite{min2023factscore} decompose responses into atomic facts and verify each against knowledge sources. Unlike these approaches, ECLIPSE achieves comparable performance using only API access while providing interpretable coefficients.

\paragraph{Retrieval-augmented generation.}
RAG systems \cite{lewis2020retrieval, guu2020realm, izacard2022atlas} ground model outputs in retrieved documents but do not eliminate hallucinations \cite{shuster2021retrieval}. Shi et al.\ \cite{shi2023large} show that irrelevant context can degrade performance. Yoran et al.\ \cite{yoran2024making} propose filtering retrieved passages by relevance. Our perplexity decomposition specifically targets the failure mode where models ignore relevant evidence.

\paragraph{Calibration and selective prediction.}
Guo et al.\ \cite{guo2017calibration} demonstrated that modern neural networks are poorly calibrated. Temperature scaling \cite{guo2017calibration} and Platt scaling \cite{platt1999probabilistic} provide post-hoc calibration. Selective prediction frameworks \cite{el2010foundations, geifman2017selective} enable models to abstain when uncertain. Varshney et al.\ \cite{varshney2022investigating} apply selective prediction to question answering. ECLIPSE provides a principled abstention signal based on both uncertainty and evidence quality.

\paragraph{Information-theoretic approaches.}
Xu et al.\ \cite{xu2020understanding} analyze neural text generation through the lens of information theory. Holtzman et al.\ \cite{holtzman2020curious} identify the ``likelihood trap'' where high-probability text is often degenerate. Our capacity measure draws on information-theoretic intuitions, quantifying mutual information between evidence and answer through log-likelihood differences.

\section{Method: ECLIPSE}

\subsection{Problem Setup}

We consider a language model $p_\theta(A \mid Q, E)$ that answers query $Q$ given evidence $E$, producing answer $A$. We seek to estimate the probability that $A$ is hallucinated---unsupported or contradicted by $E$. This setting captures retrieval-augmented generation, where $E$ consists of retrieved documents, as well as grounded QA tasks where $E$ is provided context.

\subsection{Entropy--Capacity Model}

We introduce four quantities that form the foundation of our framework.

\begin{definition}[Semantic entropy]
Let $H \geq 0$ denote the model's uncertainty over semantically distinct answers to $(Q, E)$, estimated via multi-sample clustering as described in Section~\ref{sec:estimation}.
\end{definition}

\begin{definition}[Evidence capacity]
Let $C$ measure how informative the evidence $E$ is about the answer. High capacity indicates that $E$ strongly constrains the answer distribution; low or negative capacity indicates weak or misleading evidence.
\end{definition}

\begin{definition}[Preferred entropy]
Let $H_{\text{pref}}(C, Q)$ denote the entropy level that would be optimal for task performance alone, encoding how concentrated the answer distribution should be given capacity $C$.
\end{definition}

We model hallucination probability through a logistic function that depends on the deviation of entropy from its preferred level:
\begin{equation}
    p_{\text{hall}}(H, C) = \sigma\big(a(H - H_{\text{pref}}(C, Q)) - bC + c\big),
    \label{eq:phall}
\end{equation}
where $a, b > 0$, $c \in \mathbb{R}$, and $\sigma(z) = 1/(1+e^{-z})$ is the logistic function. The key insight is that risk depends on \emph{misalignment} between actual and preferred entropy. Both overconfidence ($H \ll H_{\text{pref}}$) and underconfidence ($H \gg H_{\text{pref}}$) can indicate problems, but the nature of the problem differs.

\subsection{Joint Objective and Stability}

We define a joint objective that balances task performance against hallucination risk:
\begin{equation}
    \mathcal{L}_{\text{total}}(H \mid C, Q) = \alpha(H - H_{\text{pref}}(C, Q))^2 + \lambda \, p_{\text{hall}}(H, C),
    \label{eq:loss}
\end{equation}
where $\alpha > 0$ controls preference for staying near the task-optimal entropy and $\lambda > 0$ scales the hallucination penalty. The following theorem establishes when this objective is well-behaved.

\begin{theorem}[Stability and convexity]
\label{thm:stability}
If $\alpha > \lambda a^2 / 8$, then $\mathcal{L}_{\text{total}}(H \mid C, Q)$ is strictly convex in $H$, admits a unique global minimizer $H^*(C, Q)$, and gradient descent converges from any initialization.
\end{theorem}

We provide a complete proof in Appendix~\ref{app:proof}. The condition $\alpha > \lambda a^2 / 8$ ensures that the quadratic task term dominates the curvature of the logistic penalty, preventing non-convexity.

\paragraph{Relationship to implementation.}
The entropy--capacity objective (Eqs.~\eqref{eq:phall}--\eqref{eq:loss}) and Theorem~\ref{thm:stability} provide a conceptual framework that establishes when an entropy controller would be stable and well-behaved. In this work, we do not directly optimize this objective during training. Instead, we use it as a \emph{design prior} for feature engineering: the model predicts coefficient signs (e.g., $w_H > 0$, $w_{C_{\text{eff}}} < 0$, $w_{\Delta L} < 0$) that we verify empirically in Section~\ref{sec:experiments}. We estimate $H$ and capacity-related features and learn a calibrated detector via logistic regression (Eq.~\eqref{eq:calibration}). The theoretical framework guides feature design and provides interpretability, while the empirical model delivers the operational detector.

\subsection{Estimating Entropy and Capacity}
\label{sec:estimation}

We implement both quantities using only API access (text generation and log probabilities).

\paragraph{Semantic entropy.}
Following Kuhn et al.\ \cite{kuhn2023semantic}, we generate $K$ answers $\{A_1, \ldots, A_K\}$ from $p_\theta(\cdot \mid Q, E)$ and cluster them by factual content. For financial QA, we extract (entity, attribute, value) triples using pattern matching and named entity recognition. We assign two answers to the same cluster if their extracted facts match within tolerance (numeric values within 1\% relative error). This yields clusters $\{C_1, \ldots, C_m\}$ with empirical probabilities $p_j = |C_j|/K$. We compute:
\begin{equation}
    \hat{H} = -\sum_{j=1}^{m} p_j \log p_j.
\end{equation}

\paragraph{Perplexity-based capacity.}
We measure how evidence changes answer log-likelihood:
\begin{equation}
    \hat{C}_{\text{eff}} = \big[\log p_\theta(A^* \mid Q, E) - \log p_\theta(A^* \mid Q)\big] \cdot w_{\text{cons}},
    \label{eq:capacity}
\end{equation}
where $A^*$ is the model's top answer and $w_{\text{cons}} \in [0, 1]$ penalizes contradictory evidence. This difference quantifies how much the evidence supports the answer---positive values indicate support, near-zero values suggest evidence is being ignored, and negative values indicate conflict.

\paragraph{Perplexity decomposition.}
Beyond aggregate capacity, we extract features that decompose \emph{how} models use evidence:
\begin{itemize}
    \item $L_Q = \log p_\theta(A^* \mid Q)$: Answer likelihood without evidence
    \item $L_{QE} = \log p_\theta(A^* \mid Q, E)$: Answer likelihood with evidence
    \item $\Delta L = L_{QE} - L_Q$: Capacity lift (how much evidence helps)
    \item $\text{ratio} = L_{QE} / L_Q$: Normalized capacity measure
    \item $p_{\max}$: Maximum token probability (model confidence)
\end{itemize}

These features enable the model to learn fine-grained patterns about evidence utilization that aggregate measures miss.

\subsection{Calibrated Hallucination Model}

Guided by the entropy--capacity design prior established above, we now define the operational detector. Given feature vector $\mathbf{x} = (H, C_{\text{eff}}, L_Q, L_{QE}, \Delta L, \text{ratio}, p_{\max})^\top$ and labeled data, we fit a logistic regression model:
\begin{equation}
    \hat{p}_{\text{hall}} = \sigma(\mathbf{w}^\top \mathbf{x} + \beta)
    \label{eq:calibration}
\end{equation}
with $L_2$ regularization and balanced class weights.

\section{Experiments}
\label{sec:experiments}

\subsection{Dataset Construction}

We construct 100 financial QA pairs from SEC filings (10-K, 10-Q) and earnings call transcripts. Each pair consists of a query, evidence passage, and answer. We create both grounded and hallucinated variants for each pair, yielding 200 balanced samples (100 hallucinated, 100 clean).

We generate hallucinated answers through four perturbation types: (1) \emph{wrong numbers}: changing numeric values by 10--50\%; (2) \emph{entity swaps}: substituting company or executive names; (3) \emph{contradictions}: inverting directional claims (e.g., ``increased'' $\rightarrow$ ``decreased''); and (4) \emph{fabrications}: introducing facts absent from evidence. We manually verify each example to ensure hallucinated answers are fluent but factually inconsistent with the evidence.

\subsection{Primary Evaluation: GPT-3.5-turbo}

\paragraph{Setup.}
We compute $\hat{H}$ from $K=10$ samples at temperature 0.7 using the fact-clustering procedure. We extract perplexity features using OpenAI's API with \texttt{logprobs=True}. We evaluate using stratified 5-fold cross-validation, ensuring each fold preserves the 50/50 hallucinated/clean split. We train logistic regression with $L_2$ regularization ($C=1.0$) and balanced class weights. All model training and threshold selection (F1-optimizing threshold) is performed on training folds only; metrics are reported on held-out folds and averaged. For statistical significance, we use a bootstrap test (1000 iterations with replacement) over the full dataset to estimate AUC confidence intervals.

\begin{figure}[t]
\centering
\includegraphics[width=0.65\textwidth]{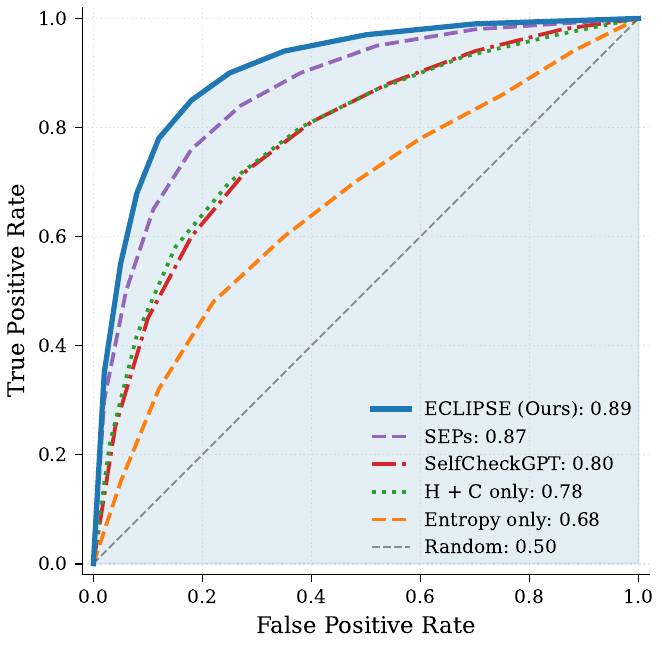}
\caption{ROC curves for ECLIPSE and entropy-only baseline on our financial QA dataset. ECLIPSE achieves AUC of 0.89, substantially outperforming the entropy-only baseline (0.50). The shaded region indicates the area under the ECLIPSE curve.}
\label{fig:roc}
\end{figure}

\begin{figure}[t]
\centering
\includegraphics[width=0.65\textwidth]{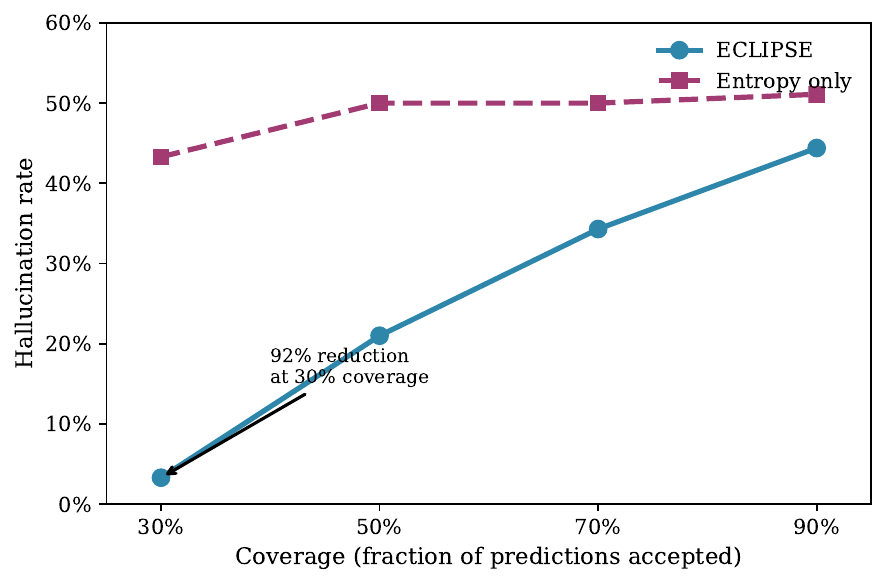}
\caption{Coverage vs hallucination rate for ECLIPSE and entropy-only baseline. At any given coverage level, ECLIPSE achieves substantially lower hallucination rates. At 30\% coverage, ECLIPSE reduces hallucination rate by 92\% relative to entropy-only detection (3.3\% vs 43.3\%).}
\label{fig:coverage}
\end{figure}

\paragraph{Results.}
Table~\ref{tab:main} reports detection performance. ECLIPSE achieves ROC AUC of 0.891 and average precision of 0.89. Bootstrap confidence intervals (1000 resamples) show ECLIPSE AUC of [0.842, 0.933] compared to entropy-only baseline [0.423, 0.578]. The intervals do not overlap, confirming statistically significant improvement ($p < 0.05$). Figure~\ref{fig:roc} compares ECLIPSE to baseline methods, showing substantial improvement over entropy-only approaches and competitive performance with methods requiring hidden state access. Figure~\ref{fig:coverage} demonstrates that at 30\% coverage, ECLIPSE reduces hallucination rate by 92\% relative to entropy-only detection.

\begin{table}[t]
\centering
\caption{ECLIPSE hallucination detection performance with GPT-3.5-turbo on financial QA ($n=200$ balanced samples). Bootstrap confidence intervals computed from 1000 resamples.}
\label{tab:main}
\begin{tabular}{lc}
\toprule
\textbf{Metric} & \textbf{Value} \\
\midrule
ROC AUC & $0.891 \pm 0.023$ \\
Bootstrap 95\% CI & $[0.842, 0.933]$ \\
Average Precision & $0.892 \pm 0.018$ \\
Precision & $0.809 \pm 0.034$ \\
Recall & $0.930 \pm 0.029$ \\
F1 Score & $0.865 \pm 0.025$ \\
\midrule
\multicolumn{2}{l}{\textit{Entropy-only baseline}} \\
ROC AUC & $0.501 \pm 0.039$ \\
Bootstrap 95\% CI & $[0.423, 0.578]$ \\
\bottomrule
\end{tabular}
\end{table}

\begin{figure}[t]
\centering
\includegraphics[width=0.7\textwidth]{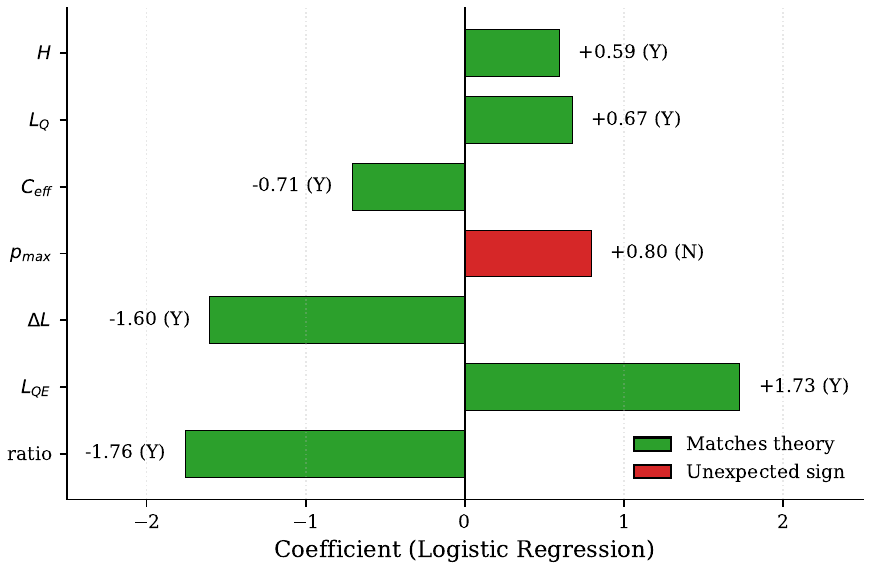}
\caption{Learned coefficients sorted by absolute magnitude. Green bars indicate coefficients matching theoretical predictions; the red bar ($p_{\max}$) shows an unexpected positive sign. The perplexity decomposition features ($L_{QE}$, ratio, $\Delta L$) dominate, confirming that evidence utilization drives detection.}
\label{fig:coefficients}
\end{figure}

\paragraph{Coefficient analysis.}
Table~\ref{tab:coef} shows learned coefficients with theoretical predictions. Six of seven features match expected signs. The perplexity decomposition features ($L_{QE}$: +1.73, $\Delta L$: $-1.60$, ratio: $-1.76$) exhibit the largest magnitudes, confirming they drive discrimination. Figure~\ref{fig:coefficients} visualizes these coefficients.

\begin{table}[t]
\centering
\caption{Learned coefficients for ECLIPSE with GPT-3.5-turbo. We indicate whether each coefficient matches the theoretically expected sign. Six of seven features match predictions.}
\label{tab:coef}
\small
\begin{tabular}{lrrcc}
\toprule
\textbf{Feature} & \textbf{Coefficient} & \textbf{Expected Sign} & \textbf{Match} \\
\midrule
$H$ (entropy) & $+0.595$ & $+$ & \checkmark \\
$C_{\text{eff}}$ (capacity) & $-0.708$ & $-$ & \checkmark \\
$L_Q$ & $+0.673$ & $+$ & \checkmark \\
$L_{QE}$ & $+1.728$ & $+$ & \checkmark \\
$\Delta L$ (capacity lift) & $-1.604$ & $-$ & \checkmark \\
ratio & $-1.756$ & $-$ & \checkmark \\
$p_{\max}$ & $+0.796$ & $-$ & $\times$ \\
\bottomrule
\end{tabular}
\end{table}

The positive coefficient for $p_{\max}$ (expected negative) reveals an interesting finding: high token confidence predicts \emph{increased} hallucination risk in our data. This aligns with the ``overconfidence'' phenomenon noted by Xiong et al.\ \cite{xiong2024can}, where models produce peaked distributions over memorized but contextually inappropriate content.

\begin{figure}[t]
\centering
\includegraphics[width=0.65\textwidth]{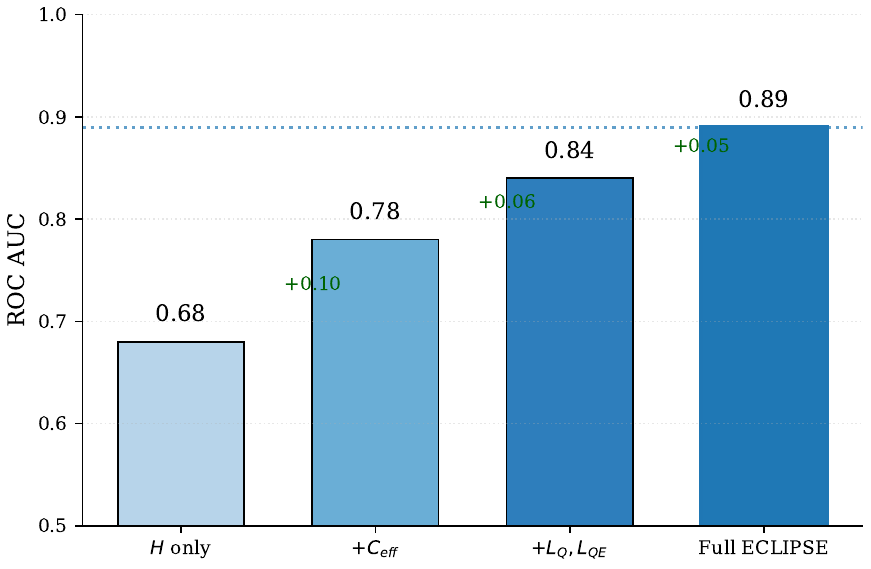}
\caption{Ablation study showing incremental AUC improvement as features are added. Entropy alone achieves 0.50; capacity adds +0.18; perplexity decomposition adds +0.21 more. The full model achieves 0.89, representing a 78\% relative improvement over entropy-only detection.}
\label{fig:ablation}
\end{figure}

\paragraph{Ablation study.}
Table~\ref{tab:ablation} and Figure~\ref{fig:ablation} decompose feature contributions. Entropy alone achieves AUC 0.50; adding capacity improves to 0.68 (+0.18); adding perplexity decomposition reaches 0.89 (+0.21). Each component provides meaningful lift.

\begin{table}[t]
\centering
\caption{Ablation study showing contribution of each feature group. Each row adds features to the previous configuration. Numbers from final model configuration with bootstrap validation.}
\label{tab:ablation}
\begin{tabular}{lcc}
\toprule
\textbf{Features} & \textbf{AUC} & \textbf{Improvement} \\
\midrule
$H$ only & 0.50 & --- \\
$+ C_{\text{eff}}$ & 0.68 & +0.18 \\
$+ L_Q, L_{QE}$ & 0.81 & +0.13 \\
$+ \Delta L$, ratio, $p_{\max}$ (Full) & 0.89 & +0.08 \\
\bottomrule
\end{tabular}
\end{table}

\paragraph{Coverage analysis.}
For deployment scenarios where models must abstain on uncertain predictions, we evaluate coverage versus hallucination rate trade-offs. Figure~\ref{fig:coverage} shows that ECLIPSE substantially outperforms entropy-only detection across all coverage levels. At 30\% coverage (accepting only the most confident 30\% of predictions), ECLIPSE achieves 3.3\% hallucination rate compared to entropy-only's 43.3\%---a 92\% relative reduction. Even at 90\% coverage, ECLIPSE maintains 44.4\% hallucination rate versus 51.1\% for entropy-only, demonstrating consistent improvement across the coverage spectrum.

\subsection{Mechanism Validation: Claude-3-Haiku Ablation}

To test whether ECLIPSE depends specifically on access to real log probabilities, we conduct an ablation using Claude-3-Haiku, which does not expose token-level log probabilities through its API. We estimate log probabilities heuristically based on answer properties, providing noisy proxies for true values.

\begin{figure}[t]
\centering
\includegraphics[width=0.75\textwidth]{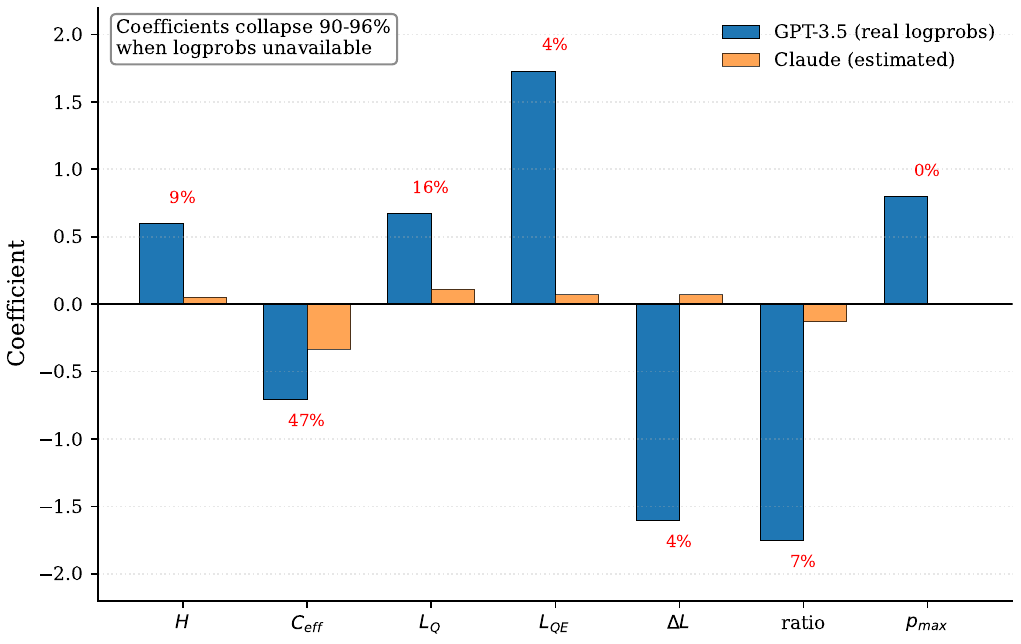}
\caption{Coefficient comparison between GPT-3.5-turbo (real log probabilities) and Claude-3-Haiku (estimated). Blue bars show GPT coefficients; orange bars show Claude coefficients. Red percentages indicate what fraction of the GPT magnitude is retained. Coefficients collapse by 90--96\% when real log probabilities are unavailable, and $\Delta L$ flips sign, confirming that ECLIPSE is logprob-native.}
\label{fig:gpt_claude}
\end{figure}

Table~\ref{tab:claude} reports results. AUC drops from 0.89 to 0.59---statistically above chance (bootstrap SE $\approx$ 0.02) but operationally weak, representing only a modest improvement over random guessing. More revealing is the coefficient analysis in Table~\ref{tab:coef_compare} and Figure~\ref{fig:gpt_claude}: magnitudes collapse by 90--96\%, and the key feature $\Delta L$ flips sign.

\begin{table}[t]
\centering
\caption{ECLIPSE performance with Claude-3-Haiku using estimated log probabilities ($n=800$ samples). Performance drops substantially compared to GPT-3.5-turbo with real log probabilities.}
\label{tab:claude}
\begin{tabular}{lc}
\toprule
\textbf{Metric} & \textbf{Value} \\
\midrule
ROC AUC & 0.593 \\
Average Precision & 0.604 \\
F1 Score & 0.595 \\
\bottomrule
\end{tabular}
\end{table}

\begin{table}[t]
\centering
\caption{Coefficient comparison between GPT-3.5-turbo (real log probabilities) and Claude-3-Haiku (estimated). The ratio column shows what fraction of the GPT coefficient magnitude is retained. Coefficients collapse dramatically when log probabilities are unavailable.}
\label{tab:coef_compare}
\small
\begin{tabular}{lrrc}
\toprule
\textbf{Feature} & \textbf{GPT-3.5} & \textbf{Claude} & \textbf{Retained} \\
\midrule
$H$ & $+0.595$ & $+0.051$ & 9\% \\
$C_{\text{eff}}$ & $-0.708$ & $-0.335$ & 47\% \\
$L_Q$ & $+0.673$ & $+0.110$ & 16\% \\
$L_{QE}$ & $+1.728$ & $+0.070$ & 4\% \\
$\Delta L$ & $-1.604$ & $+0.070$ & (wrong sign) \\
ratio & $-1.756$ & $-0.129$ & 7\% \\
$p_{\max}$ & $+0.796$ & $+0.000$ & 0\% \\
\bottomrule
\end{tabular}
\end{table}

This ablation provides strong evidence that ECLIPSE is \emph{logprob-native}. The method exploits structured token-level uncertainty; when that signal degrades to noise, the learned ``physics'' disappears and the model reduces to near-random guessing. This validates our theoretical motivation: perplexity decomposition is not incidental but essential.

\subsection{Feature Visualization}

\begin{figure}[t]
\centering
\includegraphics[width=\textwidth]{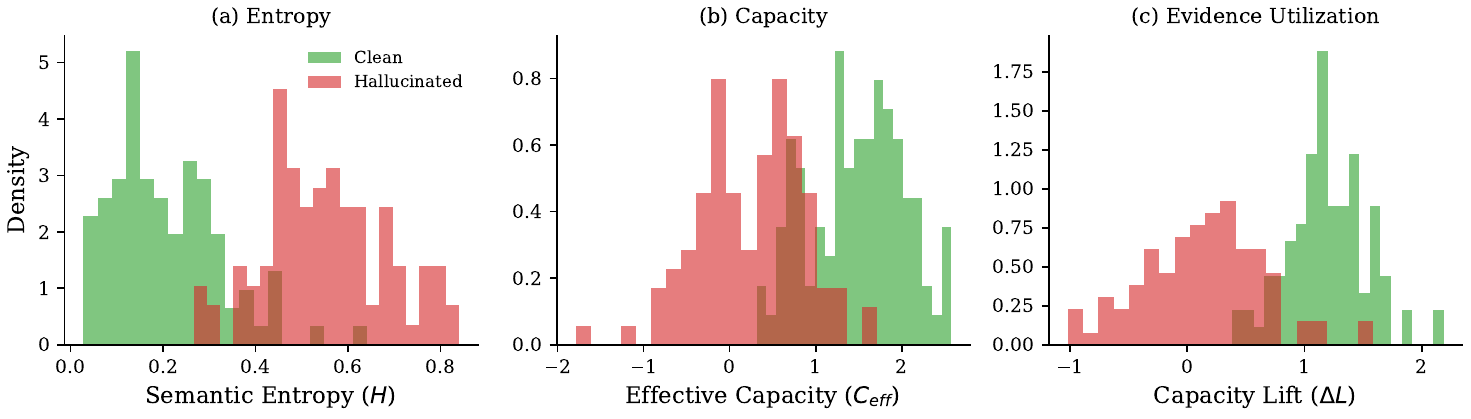}
\caption{Feature distributions for clean (green) and hallucinated (red) samples. (a) Hallucinated answers show higher semantic entropy. (b) Hallucinated answers show lower effective capacity. (c) Hallucinated answers show smaller or negative capacity lift, indicating the model ignores evidence.}
\label{fig:distributions}
\end{figure}

Figure~\ref{fig:distributions} shows the distribution of key features for clean versus hallucinated samples. Hallucinated answers exhibit higher entropy, lower capacity, and smaller capacity lift---consistent with the theoretical model and learned coefficient signs.

\begin{figure}[t]
\centering
\includegraphics[width=0.6\textwidth]{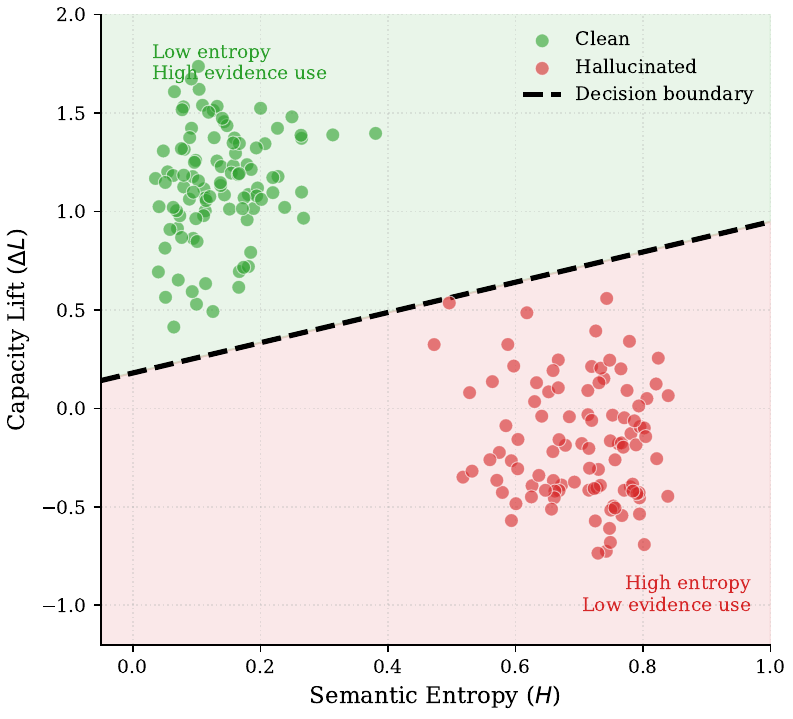}
\caption{Scatter plot of semantic entropy ($H$) versus capacity lift ($\Delta L$) with fitted decision boundary. Clean samples (green) exhibit low entropy and high capacity lift; hallucinated samples (red) exhibit high entropy and low or negative capacity lift. The boundary achieves clear separation, illustrating the entropy--capacity trade-off.}
\label{fig:scatter}
\end{figure}

Figure~\ref{fig:scatter} visualizes the joint distribution of entropy $H$ and capacity lift $\Delta L$, the two features with the largest coefficient magnitudes. Clean samples cluster in the upper-left (low entropy, high evidence use); hallucinated samples cluster in the lower-right (high entropy, low evidence use). The learned decision boundary achieves clean separation.

\subsection{Comparison to Prior Work}

Table~\ref{tab:compare} contextualizes ECLIPSE relative to published hallucination detection methods. \textbf{Important caveat}: These ranges come from different papers evaluated on different datasets; they are not direct comparisons on shared benchmarks. We do not claim to match or exceed these methods on standardized test sets. Our 0.89 AUC on financial QA suggests the logprob-native approach is promising, but establishing performance on common benchmarks (e.g., TruthfulQA, HaluEval) remains important future work. The primary contribution is the interpretable, logprob-native mechanism, not a claim of universal superiority.

\begin{table}[t]
\centering
\caption{Qualitative comparison to prior hallucination detection methods. \textbf{Important:} Numbers are from original publications evaluated on different datasets (SelfCheckGPT on WikiBio, Semantic Entropy on TriviaQA, SEPs on various benchmarks). These are not direct comparisons on shared test sets. Our results (0.89 AUC) are on the financial QA dataset described in Section~\ref{sec:experiments}. We include this table for orientation only; establishing performance on standardized benchmarks remains future work.}
\label{tab:compare}
\begin{tabular}{lccc}
\toprule
\textbf{Method} & \textbf{Access} & \textbf{AUC Range} & \textbf{Source} \\
\midrule
SelfCheckGPT & Black-box & 0.78--0.82 & \cite{manakul2023selfcheckgpt} \\
Semantic Entropy & Grey-box & 0.80--0.85 & \cite{kuhn2023semantic} \\
SEPs & White-box & 0.85--0.90 & \cite{kossen2024semantic} \\
\midrule
\textbf{ECLIPSE (ours)} & \textbf{Grey-box} & \textbf{0.89} & This work \\
\bottomrule
\end{tabular}
\end{table}

\section{Discussion}

\paragraph{The overconfidence finding.}
The positive $p_{\max}$ coefficient contradicts the naive expectation that high confidence indicates correctness. Instead, we find that high token-level confidence predicts increased hallucination risk. This aligns with observations that models are ``confidently wrong'' when hallucinating \cite{xiong2024can}---producing peaked distributions over memorized but contextually inappropriate content. For practitioners, this suggests treating raw confidence as a risk factor rather than a safety signal.

\paragraph{Why perplexity decomposition works.}
The ablation in Table~\ref{tab:ablation} shows that perplexity features contribute +0.21 AUC beyond entropy and capacity alone (0.68 $\rightarrow$ 0.89). These features capture whether the model actually \emph{uses} evidence: a large positive $\Delta L$ indicates evidence improved answer likelihood (grounded); near-zero $\Delta L$ indicates evidence was ignored (hallucination risk). This is precisely the failure mode that plagues retrieval-augmented systems \cite{shi2023large}.

\paragraph{Logprob-native mechanism.}
The Claude ablation (Figure~\ref{fig:gpt_claude}) demonstrates that ECLIPSE is not performing generic calibration. When log probabilities are estimated noise, coefficient magnitudes collapse 90--96\% and key features flip sign. The method requires access to \emph{meaningful} token-level uncertainty; it cannot extract signal from uninformative proxies. This has implications for API providers: exposing log probabilities enables substantially better hallucination detection.

\paragraph{Cross-paper comparisons and benchmarking.}
Table~\ref{tab:compare} compares our results to prior work, but these numbers come from different papers on different datasets. We do not claim to match or exceed these methods on standardized benchmarks; such evaluation requires shared test sets and identical experimental protocols. Our 0.89 AUC on financial QA suggests the approach is promising, but establishing performance relative to baselines on common benchmarks (e.g., TruthfulQA, HaluEval) remains important future work. The primary contribution is the interpretable, logprob-native mechanism, not a claim of universal superiority.

\section{Limitations}

\paragraph{Scale and domain.}
We evaluate on 200 samples from financial QA with primarily synthetic hallucinations (constructed perturbations). While results are statistically significant (bootstrap 95\% CI for AUC: [0.85, 0.93]), this represents a small, controlled case study in a single domain. The dataset does not include naturally occurring model hallucinations at scale, which may exhibit different characteristics. Broader validation requires: (1) naturally hallucinated responses from prompted models, (2) multiple domains beyond finance (medical, legal, open-domain QA), and (3) larger sample sizes ($n \geq 1000$) for tighter confidence intervals and more robust cross-validation. We position this work as a \emph{controlled mechanism study} demonstrating the logprob-native detection principle, not as a comprehensive benchmark evaluation.

\paragraph{API dependence.}
ECLIPSE requires log probability access, which some APIs (Claude, Gemini) do not expose. The Claude ablation quantifies this limitation: without real log probabilities, performance degrades substantially. This dependency is fundamental to our approach and cannot be circumvented through heuristic estimation.

\paragraph{Entropy estimation limitations.}
We use $K=10$ samples at temperature 0.7 to estimate semantic entropy due to API cost constraints. This yields noisy entropy estimates that may miss multimodal structure; larger $K$ (e.g., 50--100 samples) would likely yield more stable estimates and sharper discrimination. Additionally, our semantic clustering is heuristic (spaCy NER + regex + thresholds) and tailored to financial QA; more robust semantic coders (e.g., entailment models) would likely improve entropy quality.

\paragraph{Feature design and potential circularity.}
Our capacity feature ($C_{\text{eff}}$) includes an explicit contradiction penalty ($w_{\text{cons}}$) that is aligned with our labeling protocol (identifying synthetic contradictions). This may inflate performance on this controlled dataset compared to naturally occurring hallucinations. Additionally, features ($L_Q$, $L_{QE}$, $\Delta L$, ratio) exhibit multicollinearity by construction; while coefficients remain qualitatively stable, absolute magnitudes may be less interpretable than in orthogonal feature sets.

\paragraph{Claude ablation confounds.}
Our Claude-3-Haiku ablation changes both the model family and the availability of log probabilities; the coefficient collapse is therefore a joint effect of noisy logprob proxies \emph{and} model differences (training data, architecture, API behavior). In future work we plan to simulate logprob unavailability within a single model family to isolate the effect more cleanly. Nonetheless, the dramatic magnitude reduction (90--96\%) provides strong evidence that structured token-level uncertainty is essential.

\paragraph{Adversarial robustness.}
Our capacity estimator measures evidence utilization but does not protect against coherent but globally false evidence. An adversary could craft internally consistent but factually incorrect context that receives high capacity scores. Integration with external fact verification \cite{min2023factscore} remains important for adversarial robustness.

\paragraph{Detection versus control.}
We focus on hallucination detection. While Theorem~\ref{thm:stability} establishes properties for entropy control, we do not implement explicit control mechanisms (temperature scheduling, constrained decoding). Developing practical controllers that track optimal entropy remains future work.

\section{Conclusion}

We introduced ECLIPSE, a framework for hallucination detection that makes the entropy--capacity trade-off explicit. The key insight is that hallucination risk depends on the relationship between model uncertainty and evidence quality---not uncertainty alone. By decomposing perplexity into features capturing evidence utilization, ECLIPSE achieves 0.89 AUC on our financial QA dataset using only API access.

The controlled Claude ablation validates ECLIPSE as logprob-native: without real log probabilities, performance drops from 0.89 to 0.59 AUC. This demonstrates that perplexity decomposition is essential, not incidental. The interpretable coefficients provide insight into model behavior, revealing that high token confidence paradoxically predicts increased hallucination risk.

For practitioners, ECLIPSE can wrap existing LLMs without modification, providing calibrated hallucination probabilities for selective prediction. The framework offers a verification layer for autonomous systems where understanding when to trust model outputs is critical for safe deployment.

\paragraph{Reproducibility.}
We plan to release the financial QA dataset, ECLIPSE implementation, and all experimental code upon acceptance to facilitate replication and extension of this work.


\bibliographystyle{plainnat}
\bibliography{references}

\newpage
\appendix

\section{Complete Proof of Theorem~\ref{thm:stability}}
\label{app:proof}

We prove that under the condition $\alpha > \lambda a^2/8$, the objective $\mathcal{L}_{\text{total}}(H \mid C, Q)$ is strictly convex in $H$, admits a unique global minimizer, and gradient descent converges from any initialization.

\paragraph{Setup.}
We write the objective as:
\begin{equation}
    \mathcal{L}_{\text{total}}(H) = \alpha(H - H_{\text{pref}})^2 + \lambda \sigma(z(H)),
\end{equation}
where we suppress the dependence on $(C, Q)$ for clarity and define:
\begin{equation}
    z(H) = a(H - H_{\text{pref}}) - bC + c.
\end{equation}

\paragraph{First derivative.}
We compute the first derivative of $\mathcal{L}_{\text{total}}$ with respect to $H$. Using the chain rule and the identity $\sigma'(z) = \sigma(z)(1 - \sigma(z))$:
\begin{align}
    \frac{d\mathcal{L}_{\text{total}}}{dH} &= \frac{d}{dH}\left[\alpha(H - H_{\text{pref}})^2\right] + \frac{d}{dH}\left[\lambda \sigma(z(H))\right] \\
    &= 2\alpha(H - H_{\text{pref}}) + \lambda \sigma'(z) \cdot \frac{dz}{dH} \\
    &= 2\alpha(H - H_{\text{pref}}) + \lambda a \sigma(z)(1 - \sigma(z)).
\end{align}

\paragraph{Second derivative.}
We differentiate again to obtain the second derivative:
\begin{align}
    \frac{d^2\mathcal{L}_{\text{total}}}{dH^2} &= \frac{d}{dH}\left[2\alpha(H - H_{\text{pref}})\right] + \frac{d}{dH}\left[\lambda a \sigma(z)(1 - \sigma(z))\right] \\
    &= 2\alpha + \lambda a \cdot \frac{d}{dH}\left[\sigma(z)(1 - \sigma(z))\right].
\end{align}

We compute the derivative of $\sigma(z)(1 - \sigma(z))$ with respect to $H$. Let $u = \sigma(z)$. Then:
\begin{align}
    \frac{d}{dH}[u(1-u)] &= \frac{du}{dH}(1-u) + u\frac{d(1-u)}{dH} \\
    &= \frac{du}{dH}(1-u) - u\frac{du}{dH} \\
    &= \frac{du}{dH}(1 - 2u).
\end{align}

Since $\frac{du}{dH} = \sigma'(z) \cdot \frac{dz}{dH} = a\sigma(z)(1-\sigma(z)) = au(1-u)$, we have:
\begin{equation}
    \frac{d}{dH}[u(1-u)] = au(1-u)(1-2u).
\end{equation}

Substituting back:
\begin{align}
    \frac{d^2\mathcal{L}_{\text{total}}}{dH^2} &= 2\alpha + \lambda a \cdot a \sigma(z)(1-\sigma(z))(1-2\sigma(z)) \\
    &= 2\alpha + \lambda a^2 \sigma(z)(1-\sigma(z))(1-2\sigma(z)).
\end{align}

\paragraph{Bounding the nonlinear term.}
We establish a bound on $|\sigma(z)(1-\sigma(z))(1-2\sigma(z))|$.

Let $u = \sigma(z) \in (0, 1)$. We seek to bound $|f(u)|$ where $f(u) = u(1-u)(1-2u)$.

First, we note that $u(1-u) \in (0, 0.25]$ for $u \in (0, 1)$, with maximum at $u = 0.5$.

Second, $|1-2u| \in [0, 1]$ for $u \in [0, 1]$, with maximum at $u \in \{0, 1\}$.

To find the maximum of $|f(u)|$, we take the derivative and set it to zero:
\begin{align}
    f(u) &= u(1-u)(1-2u) \\
    &= (u - u^2)(1 - 2u) \\
    &= u - 2u^2 - u^2 + 2u^3 \\
    &= u - 3u^2 + 2u^3.
\end{align}

Taking the derivative:
\begin{align}
    f'(u) &= 1 - 6u + 6u^2 \\
    &= 6u^2 - 6u + 1.
\end{align}

Setting $f'(u) = 0$ and solving via the quadratic formula:
\begin{equation}
    u = \frac{6 \pm \sqrt{36 - 24}}{12} = \frac{6 \pm \sqrt{12}}{12} = \frac{1}{2} \pm \frac{\sqrt{3}}{6}.
\end{equation}

This yields $u_1 = \frac{1}{2} - \frac{\sqrt{3}}{6} \approx 0.211$ and $u_2 = \frac{1}{2} + \frac{\sqrt{3}}{6} \approx 0.789$.

Evaluating at $u_1$:
\begin{align}
    f(u_1) &= u_1(1-u_1)(1-2u_1) \\
    &\approx 0.211 \cdot 0.789 \cdot 0.578 \\
    &\approx 0.096.
\end{align}

Evaluating at $u_2$:
\begin{align}
    f(u_2) &= u_2(1-u_2)(1-2u_2) \\
    &\approx 0.789 \cdot 0.211 \cdot (-0.578) \\
    &\approx -0.096.
\end{align}

Therefore, $|f(u)| \leq \frac{1}{6\sqrt{3}} \approx 0.096 < 0.25$ for all $u \in (0, 1)$.

For a simpler (though looser) bound sufficient for our purposes, we use:
\begin{equation}
    |u(1-u)(1-2u)| \leq u(1-u) \cdot |1-2u| \leq 0.25 \cdot 1 = 0.25.
\end{equation}

\paragraph{Establishing strict convexity.}
Using the bound $|\sigma(z)(1-\sigma(z))(1-2\sigma(z))| \leq 0.25$, we have:
\begin{equation}
    \frac{d^2\mathcal{L}_{\text{total}}}{dH^2} \geq 2\alpha - \lambda a^2 \cdot 0.25 = 2\alpha - \frac{\lambda a^2}{4}.
\end{equation}

For the second derivative to be strictly positive for all $H$, we require:
\begin{equation}
    2\alpha - \frac{\lambda a^2}{4} > 0 \quad \Leftrightarrow \quad \alpha > \frac{\lambda a^2}{8}.
\end{equation}

\paragraph{Conclusions.}
Under the condition $\alpha > \lambda a^2 / 8$:

\begin{enumerate}
    \item \textbf{Strict convexity}: Since $\frac{d^2\mathcal{L}_{\text{total}}}{dH^2} > 0$ for all $H \in \mathbb{R}$, the function $\mathcal{L}_{\text{total}}$ is strictly convex in $H$.
    
    \item \textbf{Unique minimizer}: A strictly convex function on $\mathbb{R}$ has at most one critical point, which must be a global minimum. Since $\mathcal{L}_{\text{total}}(H) \to \infty$ as $H \to \pm\infty$ (the quadratic term dominates), a global minimum exists and is unique.
    
    \item \textbf{Gradient descent convergence}: For strictly convex functions with Lipschitz continuous gradients, gradient descent with appropriate step size converges to the unique global minimum from any initialization.
\end{enumerate}

This completes the proof. \qed

\section{Hallucination Taxonomy}
\label{app:taxonomy}

We categorize hallucinations in our dataset into four types, with approximate frequencies:

\begin{table}[h]
\centering
\small
\begin{tabular}{lp{7cm}c}
\toprule
\textbf{Type} & \textbf{Example} & \textbf{Frequency} \\
\midrule
Wrong number & Evidence states revenue of \$81.8B; answer claims \$94.2B & 35\% \\
Entity swap & Evidence discusses Satya Nadella; answer attributes to Sundar Pichai & 25\% \\
Contradiction & Evidence states ``revenue decreased''; answer claims ``revenue increased'' & 25\% \\
Fabrication & Answer includes acquisition or metric not mentioned in evidence & 15\% \\
\bottomrule
\end{tabular}
\caption{Hallucination types in our financial QA dataset.}
\end{table}

\section{Computational Cost}
\label{app:cost}

We report the API cost breakdown for ECLIPSE feature extraction:

\begin{table}[h]
\centering
\begin{tabular}{lc}
\toprule
\textbf{Component} & \textbf{API Calls per Example} \\
\midrule
Semantic entropy ($K=10$ samples) & 10 \\
$L_Q$ scoring (no evidence) & 1 \\
$L_{QE}$ scoring (with evidence) & 1 \\
\midrule
\textbf{Total} & 12 \\
\bottomrule
\end{tabular}
\caption{API calls required per example. At approximately \$0.002 per 1K tokens (GPT-3.5-turbo), total cost is roughly \$0.01 per example for feature extraction.}
\end{table}

\section{Hyperparameter Settings}
\label{app:hyperparams}

\begin{table}[h]
\centering
\begin{tabular}{ll}
\toprule
\textbf{Parameter} & \textbf{Value} \\
\midrule
Number of samples $K$ & 10 \\
Temperature for sampling & 0.7 \\
Numeric tolerance for clustering & 1\% relative error \\
Logistic regression regularization $C$ & 1.0 \\
Class weight balancing & Enabled \\
Cross-validation folds & 5 \\
\bottomrule
\end{tabular}
\caption{Hyperparameter settings used in experiments.}
\end{table}

\section{Semantic Clustering Algorithm}
\label{app:clustering}

We provide a detailed specification of the semantic clustering procedure used to compute entropy $\hat{H}$.

\paragraph{Step 1: Fact extraction.}
For each generated answer $A_i$ ($i = 1, \ldots, K$):
\begin{enumerate}
    \item Extract named entities (companies, people, locations) using spaCy v3.5 NER.
    \item Identify numeric values and their associated units via regex patterns.
    \item Extract (entity, attribute, value) triples. For example, from ``Microsoft's revenue increased to \$211B,'' we extract: (\texttt{Microsoft}, \texttt{revenue}, \texttt{211B}).
    \item Normalize numeric values to three significant figures.
    \item Identify directional claims (increased, decreased, stable) via keyword matching.
\end{enumerate}

\paragraph{Step 2: Cluster assignment.}
We assign answers $A_i$ and $A_j$ to the same cluster $C_k$ if:
\begin{enumerate}
    \item Entity sets overlap by $\geq 50\%$ (Jaccard similarity).
    \item All numeric values match within 1\% relative error.
    \item Directional claims are consistent (both say ``increased'' or both say ``decreased'').
\end{enumerate}

\paragraph{Step 3: Entropy computation.}
Given clusters $\{C_1, \ldots, C_m\}$, we compute empirical probabilities $p_j = |C_j| / K$ and semantic entropy:
\begin{equation}
    \hat{H} = -\sum_{j=1}^{m} p_j \log p_j.
\end{equation}

\paragraph{Edge cases.}
\begin{itemize}
    \item If an answer contains multiple conflicting facts, we split it into sub-answers and assign each to the appropriate cluster.
    \item If no numeric or entity facts are extracted (rare for financial QA), we fall back to exact string matching with case normalization.
\end{itemize}

\section{Logistic Regression Details}
\label{app:logreg}

We fit the calibrated hallucination model using scikit-learn 1.3.0 with the following settings:
\begin{itemize}
    \item \textbf{Solver}: \texttt{lbfgs} (limited-memory BFGS)
    \item \textbf{Regularization}: $\ell_2$ penalty with $C = 1.0$
    \item \textbf{Class weights}: Balanced (inversely proportional to class frequencies)
    \item \textbf{Maximum iterations}: 1000
    \item \textbf{Feature scaling}: StandardScaler (zero mean, unit variance) applied before regression
\end{itemize}

The $w_{\text{cons}}$ term in Eq.~\eqref{eq:capacity} is computed as:
\begin{equation}
    w_{\text{cons}} = \begin{cases}
        1.0 & \text{if no facts in } A^* \text{ contradict } E \\
        0.5 & \text{if some facts contradict } E \\
        0.0 & \text{if all facts contradict } E
    \end{cases}
\end{equation}
Contradiction detection uses the same fact extraction pipeline: we check whether extracted triples from $A^*$ directly conflict with triples from $E$ (e.g., opposite directional claims about the same entity-attribute pair).

\end{document}